\documentclass[a4paper,fleqn]{cas-dc}
\usepackage{subcaption}
\newcommand{\xmark}{\ding{56}}%
\usepackage{pifont}
\usepackage[numbers]{natbib}

\def\tsc#1{\csdef{#1}{\textsc{\lowercase{#1}}\xspace}}
\tsc{WGM}
\tsc{QE}
\tsc{EP}
\tsc{PMS}
\tsc{BEC}
\tsc{DE}

\begin{document}
\let\WriteBookmarks\relax
\def\floatpagepagefraction{1}
\def\textpagefraction{.001}
\shorttitle{TwinMixing Model}
\shortauthors{M-K Do et-al.}

\title [mode = title]{\textbf{TwinMixing:} A Shuffle-Aware Feature Interaction Model for Multi-Task Segmentation}

\author[1,2]{Minh-Khoi Do}
\ead{khoidm.19@grad.uit.edu.vn}
\cormark[1]

\author[1,2]{Huy Che}[orcid=0009-0007-7477-4702]
\ead{huycq@uit.edu.vn}
\cormark[1]

\author[1,2]{Dinh-Duy Phan}[orcid=0000-0003-0228-9648]
\ead{duypd@uit.edu.vn}

\author[1,2]{Duc-Khai Lam}[orcid=0000-0003-2711-1408]
\ead{khaild@uit.edu.vn}
\cormark[2]

\author[1,2]{Duc-Lung Vu}[orcid=0000-0002-0045-4657]
\ead{lungvd@uit.edu.vn}


\affiliation[1]{organization={University of Information Technology},
                city={Ho Chi Minh City},
                country={Vietnam}}
                
\affiliation[2]{organization={Vietnam National University},
                city={Ho Chi Minh City},
                country={Vietnam}}

\cortext[cor1]{Equal contribution}
\cortext[cor2]{Corresponding author}

\begin{abstract}
Accurate and efficient perception is essential for autonomous driving, where segmentation tasks such as drivable-area and lane segmentation provide critical cues for motion planning and control. However, achieving high segmentation accuracy while maintaining real-time performance on low-cost hardware remains a challenging problem. To address this issue, we introduce TwinMixing, a lightweight multi-task segmentation model designed explicitly for drivable-area and lane segmentation. The proposed network features a shared encoder and task-specific decoders, enabling both feature sharing and task specialization. Within the encoder, we propose an Efficient Pyramid Mixing (EPM) module that enhances multi-scale feature extraction through a combination of grouped convolutions, depthwise dilated convolutions and channel shuffle operations, effectively expanding the receptive field while minimizing computational cost. Each decoder adopts a Dual-Branch Upsampling (DBU) Block composed of a learnable transposed convolution–based \textit{Fine detailed branch} and a parameter-free bilinear interpolation–based \textit{Coarse grained branch}, achieving detailed yet spatially consistent feature reconstruction. Extensive experiments on the BDD100K dataset validate the effectiveness of TwinMixing across three configurations - \textit{tiny}, \textit{base}, and \textit{large}. Among them, the \textit{base} configuration achieves the best trade-off between accuracy and computational efficiency, reaching 92.0\% mIoU for drivable-area segmentation and 32.3\% IoU for lane segmentation with only 0.43M parameters and 3.95 GFLOPs. Moreover, TwinMixing consistently outperforms existing segmentation models on the same tasks, as illustrated in Figure \ref{fig:compare}. Thanks to its compact and modular design, TwinMixing demonstrates strong potential for real-time deployment in autonomous driving and embedded perception systems. The source code is avaiable at \href{https://github.com/Jun0se7en/TwinMixing}{https://github.com/Jun0se7en/TwinMixing}.

\end{abstract}


\maketitle

\section{Introduction}

In autonomous driving systems, image segmentation plays a vital role in understanding the driving environment by identifying key elements such as road surfaces, vehicles, pedestrians, and lane markings. However, constructing highly detailed multi-class semantic maps \cite{cityscape,synthia,ACDC,gtav} is often unnecessary for practical advanced driver-assistance systems (ADAS) \cite{adas1,adas2,adas3}. A critical requirement of ADAS is real-time perception, where even minor delays in processing can compromise driving safety. Instead of exhaustively segmenting all objects in the scene, ADAS applications primarily focus on extracting task-relevant regions-notably drivable areas and lane markings-which directly support navigation, lane keeping, and trajectory planning. Consequently, segmentation approaches that emphasize task-specific utility and real-time efficiency are more desirable than those that solely focus on fine-grained semantic labeling.

\begin{figure}
    \centering
    \includegraphics[width=\linewidth]{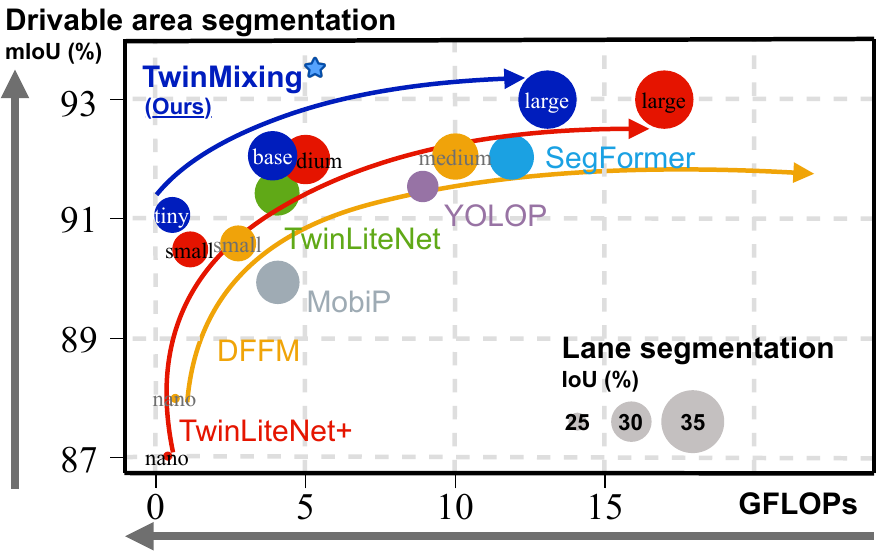}
    \caption{The horizontal axis represents FLOPs, the vertical axis denotes mIoU, and the circle radius corresponds to IoU.}
    \label{fig:compare}
\end{figure}

In contrast to general semantic scene understanding tasks \cite{cityscape,synthia,ACDC,gtav,bdd} that focus on categorizing all visible objects within a scene, drivable area segmentation and lane segmentation \cite{bdd} serve a more safety-critical function in autonomous driving perception. As illustrated in Fig.~\ref{fig:compare}, these tasks provide crucial information for identifying navigable regions, maintaining stable lane positioning, and enabling trajectory planning and collision avoidance in complex and dynamic traffic environments. The primary aim of segmentation research in this context is to create accurate mappings between sensory inputs, such as camera images, LiDAR, or other modalities, and their corresponding segmentation maps. Nevertheless, achieving high accuracy alone is not sufficient for real-world deployment; models must also deliver real-time inference performance to ensure safety and responsiveness in autonomous driving systems. This requirement underscores the need for segmentation algorithms that strike a balance between strong representational capability and computational efficiency, while maintaining high processing throughput on resource-constrained onboard hardware. Consequently, balancing accuracy and efficiency emerges as a key challenge in designing segmentation systems for practical autonomous driving applications.



Although recent segmentation studies have achieved remarkable accuracy \cite{interactive,yolopv2,yolopv3,yolopx}, most are evaluated on high-end GPUs, which differ substantially from the computational constraints of in-vehicle systems. To balance accuracy and efficiency, several lightweight architectures \cite{twin,twinplus,BILane} have been developed. Among them, dilated convolution–based models \cite{twin,twinplus} effectively enlarge the receptive field without increasing the kernel size, thereby enhancing contextual perception at a moderate cost. However, these approaches still suffer from redundant computation and limited channel interaction. To overcome these limitations, we introduce an efficient feature extraction strategy that combines dilated depthwise convolutions with channel shuffle operations. Our proposed model takes a different perspective by emphasizing inter-feature interaction and cross-scale information integration within the encoder. Instead of relying solely on structural efficiency, TwinMixing enhances representational richness through efficient pyramid mixing and channel-level fusion. This design strengthens cross-channel communication while substantially reducing parameters and FLOPs, resulting in a more compact yet expressive encoder suitable for embedded deployment.

\begin{figure}
    \centering
    \includegraphics[width=\linewidth]{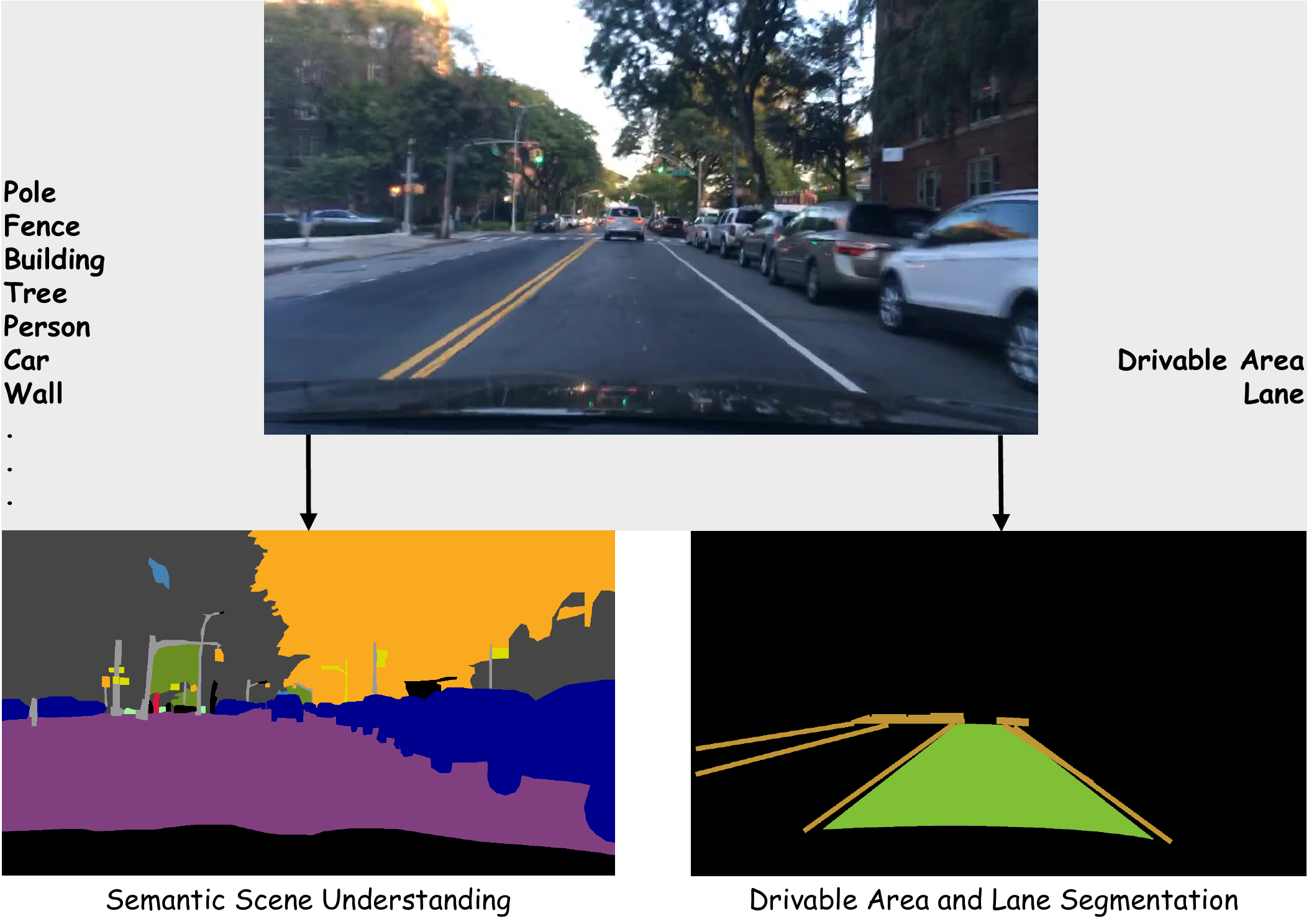}
    \caption{Visual comparison of semantic scene understanding versus drivable area and lane segmentation, highlighting the focus on safety-critical and navigable regions for autonomous driving.}
    \label{fig:placeholder}
\end{figure}

Building upon these principles, we propose the TwinMixing Model, a lightweight multi-task segmentation architecture designed for drivable area and lane segmentation. The model employs a shared encoder that extracts multi-scale features through a combination of Efficient Pyramid Mixing (EPM), followed by two task-specific decoders equipped with Dual Branch Upsampling Blocks (DBU) to reconstruct segmentation masks for each task. The integration of EPM Units enables efficient multi-scale representation learning, while the dual-branch decoder balances detailed reconstruction with spatial smoothness through the fusion of transposed convolution and bilinear upsampling. Our main contributions can be summarized as follows:

\begin{itemize}
\item We propose TwinMixing, a lightweight multi-task segmentation model that jointly performs drivable-area and lane-line segmentation. The shared encoder integrates dilated depthwise separable convolutions, channel shuffle operations, and the proposed Efficient Pyramid Mixing (EPM) module to achieve rich multi-scale representations with minimal computational overhead.

\item We design Dual Branch Upsampling (DBU)–that fuses a learnable transposed convolution–based Fine Branch with a parameter-free Bilinear Upsampling–based Coarse Branch. This dual-path strategy enhances spatial detail recovery and stability while avoiding checkerboard artifacts.         ,

\item TwinMixing is developed in three scalable configurations-\textit{tiny}, \textit{base}, and \textit{large}-ranging from 0.10M / 1.08 GFLOPs to 1.50M / 14.25 GFLOPs, supporting flexible deployment from embedded systems to edge devices.

\item Extensive experiments on the BDD100K dataset demonstrate that TwinMixing$_\text{large}$ achieves 92.8\% mIoU for drivable-area segmentation and 34.2\% IoU for lane segmentation, outperforming recent lightweight baselines such as TwinLiteNet$^+$ \cite{twinplus}, DFFM \cite{dffm}, and IALaneNet \cite{interactive} with significantly lower computational cost.
\end{itemize}

\section{\uppercase{Related Work}}

Semantic segmentation is a fundamental task in computer vision, in which each pixel in an image is assigned a semantic label. In autonomous driving, this task plays a crucial role in environmental perception and scene understanding \cite{cityscape,synthia,ACDC,gtav,bdd}, supporting downstream tasks such as motion planning, navigation, and collision avoidance. Pioneering works such as DeepLab~\cite{deeplab}, SegFormer~\cite{segformer}, and Mask2Former \cite{mask2former} have achieved high accuracy through deep feature extraction and strong contextual representation. However, these models often rely on heavy backbones with high computational costs, making them unsuitable for real-time deployment on embedded or in-vehicle systems. To improve computational efficiency, several lightweight architectures \cite{enet,FastSCNN,espnet} have been proposed, which employ simplified encoder–decoder designs to reduce complexity while maintaining reasonable segmentation performance. In addition, channel mixing mechanisms such as grouped convolution and channel shuffle, introduced in ShuffleNet~\cite{shufflenet} and MobileNet~\cite{mobilenet}, further enhance feature diversity without significantly increasing the parameter count.

In practical autonomous driving systems, however, segmenting a wide range of complex semantic categories (e.g., road, building, vegetation, pedestrian, or traffic sign) is not always necessary for real-time decision-making. Instead, advanced driver assistance systems (ADAS)  often focus on perception tasks that directly support vehicle control, such as drivable area segmentation and lane line detection. Consequently, recent studies have developed task-specific models~\cite{interactive,dffm,twin,twinplus,mobip,enet,enetsad,dltnet,pspnet} for these purposes, following either single-task \cite{enet,enetsad,dltnet,pspnet} or multi-task paradigms \cite{interactive,dffm,twin,twinplus,mobip}. Among them, multi-task models have gained more attention due to their ability to share features across tasks, reduce redundant computation, and improve scalability. Typically, these models adopt a shared encoder to learn general representations, followed by task-specific decoders to specialize for each output. Despite these advances, most existing multi-task architectures \cite{twin,twinplus,dffm} rely on dilated convolutions to enlarge the receptive field but remain limited in inter-channel interaction, which constrains feature diversity and contextual representation capacity.

\begin{figure*}
    \centering
    \includegraphics[width=\linewidth]{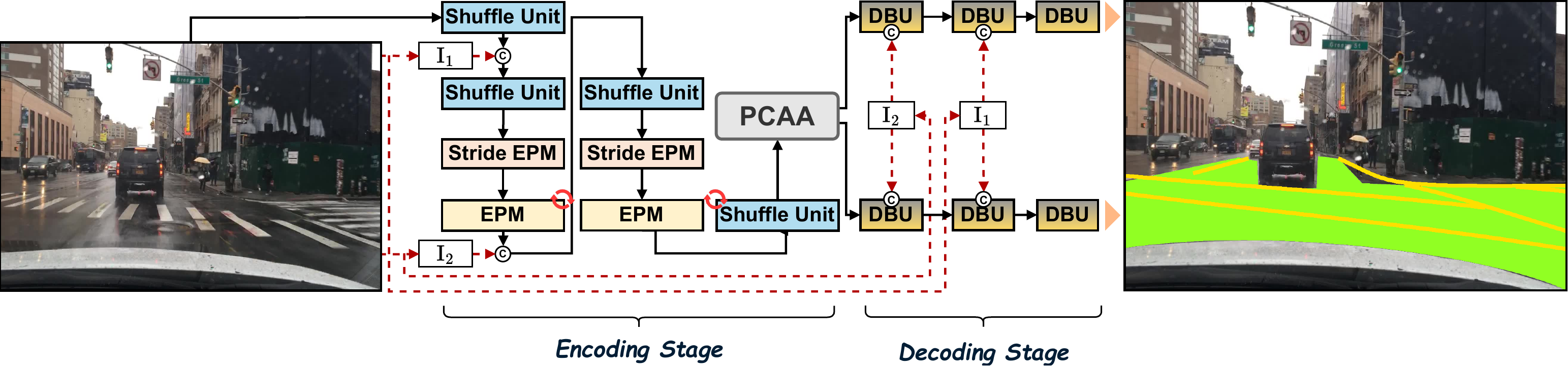}
    \caption{\textbf{The architecture of TwinMixing.} The model consists of a shared encoder and two task-specific decoders. The encoder integrates the proposed Efficient Pyramid Mixing (EPM) modules to enhance multi-scale feature extraction and contextual representation. Each decoder adopts a Dual Branch Upsampling Block (DBU) composed of \textit{Fine detailed branch} and \textit{Coarse grained branch}. The two decoders independently generate segmentation masks for lane lines and drivable areas, respectively.}
    \label{fig:arch}
\end{figure*}

From these observations, two significant challenges remain insufficiently addressed in prior research: (1) how to enhance inter-channel information exchange while maintaining effective multi-scale contextual representation under low computational cost; and (2) how to achieve stable spatial detail reconstruction during decoding without introducing boundary noise or checkerboard artifacts. To address these challenges, we propose TwinMixing - a lightweight multi-task segmentation model designed explicitly for drivable area and lane line segmentation. The model incorporates an Efficient Pyramid Mixing (EPM) module within the encoder to expand the receptive field via dilated depthwise convolutions while facilitating channel interactions through grouped convolutions and channel shuffle. In the decoding stage, a Dual Branch Upsampling (DBU) module with two parallel branches ensures smooth and stable spatial reconstruction. Through this design, TwinMixing achieves an optimal balance between computational efficiency, accuracy, and training stability, making it well-suited for real-time perception in autonomous driving systems.

\section{TwinMixing Model}

\subsection{Overall architecture}

The TwinMixing model is a multi-task segmentation architecture designed to simultaneously perform two related tasks: lane line segmentation and drivable area segmentation. The model takes an RGB image as input and extracts feature representations through a shared encoder. These shared features are then processed by two task-specific decoders, each dedicated to reconstructing a segmentation mask for its corresponding task.

The encoder of TwinMixing is built upon a combination of convolutional layers, the proposed Efficient Pyramid Mixing (EPM) modules, and the PCAA module \cite{pcaa}. This combination enhances multi-scale feature representation while maintaining high computational efficiency. Specifically, the EPM modules are designed to expand the receptive field and facilitate channel interactions at a low computational cost through multiple EPM Units, enabling the model to capture contextual information across different spatial levels. Meanwhile, the PCAA module serves as an attention mechanism that amplifies semantically important regions for each target class. During encoding, the spatial resolution of feature maps is gradually reduced from H×W to H/8×W/8 before being passed to the decoding stage.

In contrast to the shared encoder, the decoders in TwinMixing are task-specific, allowing the model to learn specialized feature representations optimized for each segmentation objective. Each decoder is constructed using the proposed Dual Branch Upsampling Block (DBU), which consists of two complementary branches: the \textit{Fine detailed branch} and the \textit{Coarse grained branch}. The outputs of these two branches are fused through element-wise addition, enabling the model to simultaneously leverage the detail reconstruction capability of transposed convolution and the spatial smoothness and stability provided by Bilinear Interpolation. As a result, TwinMixing achieves more accurate spatial feature reconstruction and improved training stability.

During inference, the encoder processes the input image $\mathcal{I}_{rgb}$ to produce an intermediate feature representation $\mathcal{F}_e$. This shared feature is then passed through two separate decoders to generate the final segmentation masks: $\mathcal{O}_{lane}$ for lane line segmentation and $\mathcal{O}_{drivable}$ for drivable area segmentation. The overall architecture of TwinMixing is illustrated in Figure \ref{fig:arch}.

\subsection{Encoder}

\begin{figure}
    \centering
    \includegraphics[width=\linewidth]{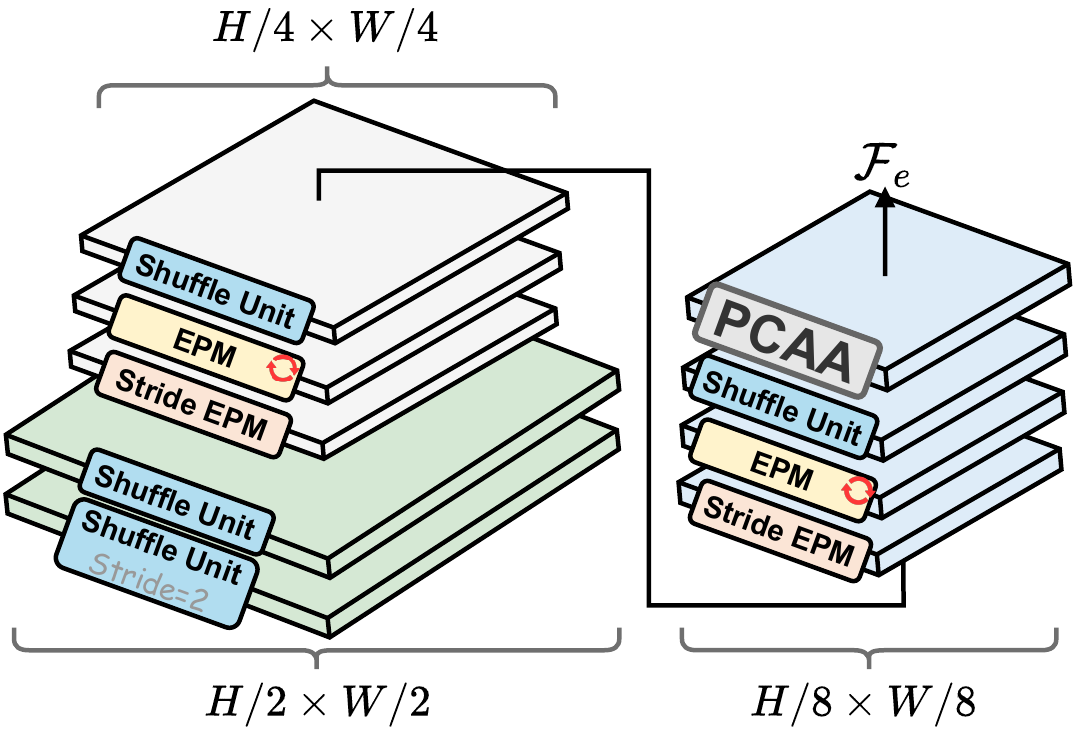}
    \caption{\textbf{Overview of the encoder in TwinMixing.} The encoder extracts hierarchical multi-scale features from the input image through a combination of standard convolutional layers, Efficient Pyramid Mixing (EPM) modules, and Partial Class Activation Attention (PCAA) \cite{pcaa}, producing the shared representation $\mathcal{F}_e$ for subsequent decoding.}
    \label{fig:arch-encoder}
\end{figure}

In the TwinMixing segmentation model, the encoding stage is implemented using a shared encoder, that extracts and represents spatial and semantic features from the input image. The encoder is constructed using a sequence of CNN layers that combine Shuffle Units \cite{shufflenet} with the proposed Efficient Pyramid Mixing (EPM) modules. The EPM modules include both EPM and Stride EPM variants. While the standard EPM focuses on expanding the receptive field to enhance multi-scale feature representation, the Stride EPM further enables the encoder to simultaneously enlarge the receptive field and perform progressive downsampling across layers. During the encoding process, the feature maps are gradually reduced in spatial resolution from H×W to H/8$\times$W/8, followed by a PCAA attention module \cite{pcaa} before being passed to the decoding stage. The encoder design is illustrated in Figure \ref{fig:arch-encoder}.

\begin{figure}
    \centering
    \includegraphics[width=\linewidth]{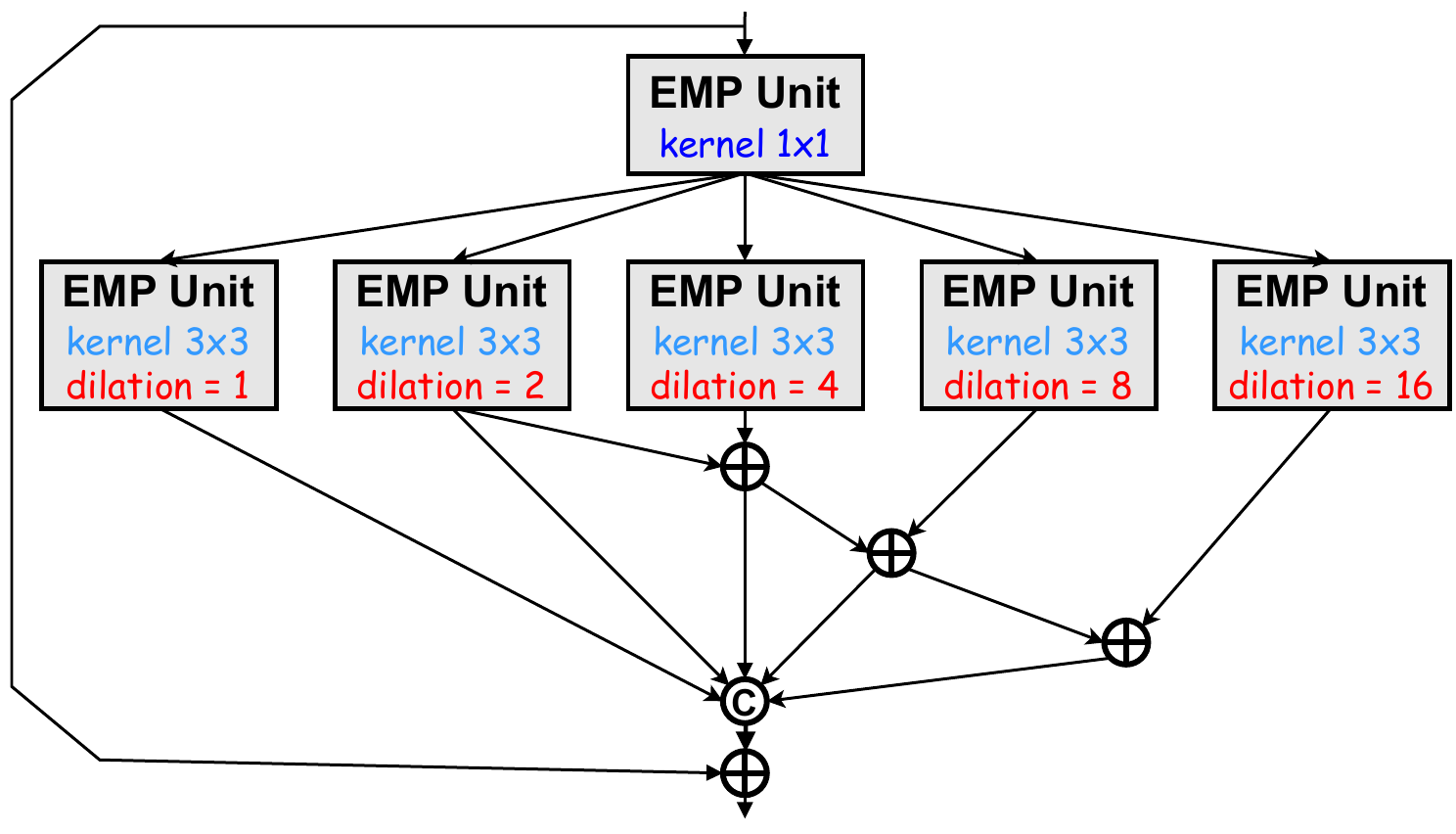}
    \caption{\textbf{Illustration of the proposed Efficient Pyramid Mixing (EPM) module.} The design is inspired by the ESP \cite{espnet}, where the EPM performs a \textit{reduction} step using an EPM Unit with a kernel 1$\times$1 before \textit{splitting} features into multiple parallel branches. Each branch \textit{transforms} the reduced feature through an EPM Unit with a different dilation rate to capture multi-scale spatial information. The outputs of all branches are then \textit{merged} through the Hierarchical Feature Fusion (HFF) mechanism \cite{espnet}. In the Stride EPM variant, the reduction step employs an EPM Unit with a kernel 1$\times$1 and a stride of 2 to achieve downsampling.}
    \label{fig:EPM}
\end{figure}

Within the encoder, the proposed Efficient Pyramid Mixing (EPM) plays a central role in enhancing multi-scale feature extraction while maintaining computational efficiency. EPM follows the \textit{reduce–split–transform–merge} principle, enabling the model to process features across multiple spatial scale representations effectively. Unlike prior approaches \cite{deeplab} that directly split the input features into multiple branches, the proposed EPM first performs dimensionality reduction through an EPM Unit with a 1$\times$1 kernel, effectively lowering the computational cost of subsequent multi-branch transformations while preserving essential representational information. By performing the reduction before branching, the subsequent transformations in each parallel path operate on lower-dimensional feature maps, thereby significantly decreasing the overall computational burden of multi-branch processing. The reduced feature is then split into multiple parallel branches, each transformed by an EPM Unit with a different dilation rate, enabling the model to capture contextual information at multiple spatial scales. EPM Unit enables the model to capture contextual information at varying receptive scales. The outputs from these branches are then fused via a Hierarchical Feature Fusion (HFF) mechanism \cite{espnet}, which mitigates gridding artifacts. The overall architecture of the EPM module is illustrated in Figure \ref{fig:EPM}.

\begin{figure*}[!b]
  \centering
  \begin{subfigure}{0.243\columnwidth}
      \includegraphics[width=\textwidth]{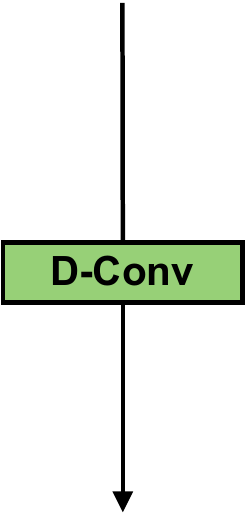}
      \caption{Dilated Convolution\\
                \textcolor{white}{a}}
      \label{esp1}
  \end{subfigure}
  \hspace{0.09\textwidth} 
  \begin{subfigure}{0.245\columnwidth}
      \includegraphics[width=\textwidth]{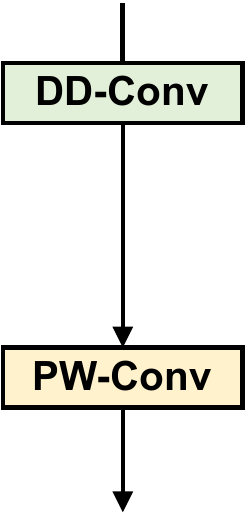}
      \caption{Depth-wise Dilated Separable Convolution}
      \label{esp2}
  \end{subfigure}
  \hspace{0.09\textwidth} 
  \begin{subfigure}{0.34\columnwidth}
      \includegraphics[width=\textwidth]{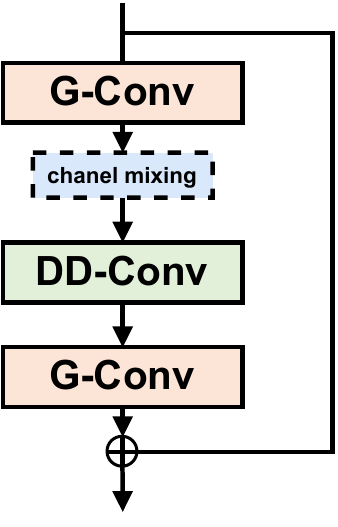}
      \caption{EPM Unit\\
                \textcolor{white}{a}\\
                \textcolor{white}{a}}
      \label{esp3}
  \end{subfigure}
  \hspace{0.09\textwidth} 
    \begin{subfigure}{0.44\columnwidth}
      \includegraphics[width=\textwidth]{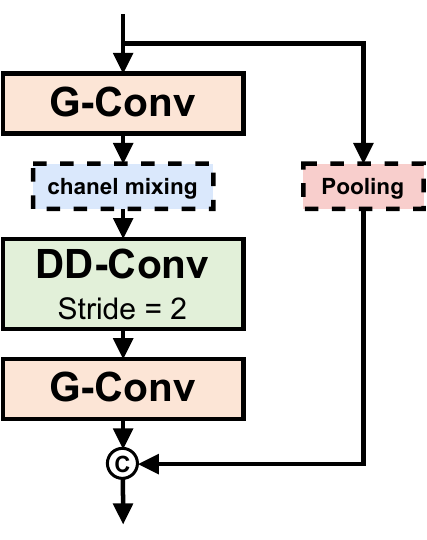}
      \caption{Stride EPM Unit\\
                \textcolor{white}{a}\\
                \textcolor{white}{a}}
      \label{esp4}
  \end{subfigure}
  \caption{\textbf{Evolution of transformation designs in approaches derived from the Efficient Spatial Pyramid (ESP) framework \cite{espnet}.} (a) Dilated Convolution (D-Conv) in ESPNet \cite{espnet}; (b) Depth-wise Dilated Separable Convolution (DDS-Conv) in \cite{twinplus,trilitenet}; (c) EPM Unit proposed; (d) Stride EPM Unit variant for downsampling.}
  \label{fig:esp_comparison}
\end{figure*}

Within the EPM module, each EPM Unit serves as the core transformation process, responsible for capturing and encoding multi-scale spatial information. Structurally, the EPM Unit is structurally optimized to achieve a balance between computational efficiency and representational capacity, combining essential operations: (i) a grouped 1×1 convolution for feature projection; (ii) a 3×3 depthwise dilated convolution for spatial transformation and receptive field expansion without incurring significant computational overhead; and (iii) A channel shuffle operation is employed to restore inter-group information flow and mitigate the isolation caused by grouped convolutions. This combination enables an effective balance between representation capability and efficiency, making the EPM Unit particularly suitable for real-time segmentation tasks that demand both high accuracy and fast inference.

In the original Efficient Spatial Pyramid (ESP) design \cite{espnet}, only dilated convolutions (D-Conv) were employed for feature transformation, as illustrated in Figure \ref{esp1}. The proposed Efficient Pyramid Mixing (EPM) Unit draws inspiration from the Depthwise Dilated Separable Convolution (DDS-Conv) structure introduced in previous studies \cite{twinplus,trilitenet}, depicted in Figure \ref{esp2}. DDS-Conv effectively enlarges the receptive field by applying a Depthwise Dilated Convolution (DD-Conv) to capture spatial context, followed by a Pointwise Convolution (PW-Conv) to adjust the channel dimensionality and restore feature compactness. However, since both operations process channels independently, DDS-Conv lacks inter-channel interaction, resulting in limited feature diversity and weak contextual representation. To overcome this limitation, the EPM Unit replaces standard PW-Convs with grouped 1$\times$1 convolutions, followed by a channel shuffle operation to promote cross-group information exchange. The shuffled features are then transformed by a DD-Conv, where the dilation rate is adaptively adjusted to control the receptive field based on network depth. Finally, a second grouped 1$\times$1 convolution restores the output channel dimension. When the number of input and output channels is identical, a shortcut connection with element-wise addition is applied to stabilize training and facilitate efficient information propagation across layers, as shown in Figure \ref{esp3}. For the Stride EPM variant, used in the reduction stages of the EPM hierarchy, a 1$\times$1 convolution with stride = 2 is employed, as illustrated in Figure \ref{esp4}. In addition to performing DD-Conv with a stride of 2, we introduce an additional 3$\times$3 average pooling layer in the shortcut branch to downsample spatially, and replace element-wise addition with channel concatenation, allowing the network to increase its output channel capacity with minimal computational overhead. This design maintains a balance between spatial efficiency and information preservation, ensuring that the downsampling process does not compromise the geometric structure of the learned feature representations.

\subsection{Decoder}

\begin{figure}
    \centering
    \includegraphics[width=\linewidth]{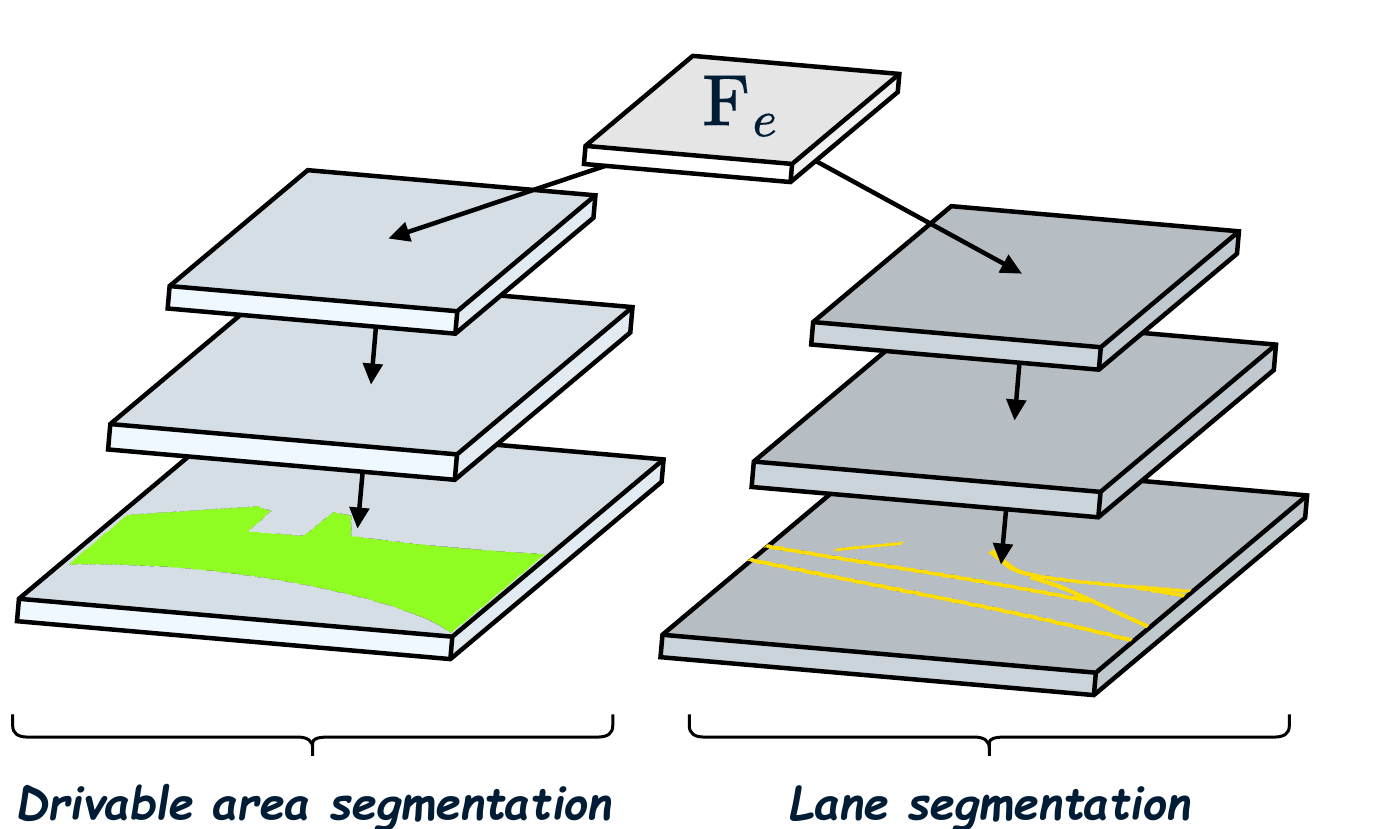}
    \caption{Architecture of the proposed model with two parallel decoders sharing the same structural design for different segmentation tasks.}
    \label{fig:dual_decoder}
\end{figure}

In the TwinMixing Model, we design separate decoders for each segmentation task to enable the model to learn task-specific feature representations. Each decoder is responsible for transforming the encoder output features ($\text{F}_e$) into a segmentation mask that has the exact spatial resolution as the input image, as illustrated in Figure \ref{fig:dual_decoder}. The decoders are constructed from a combination of convolutional operations and upsampling methods, jointly extracting spatial features and restoring resolution. In particular, we propose the Dual Branch Upsampling Block (DBU) - a dual-path upsampling structure composed of two complementary branches: the \textit{Fine detailed branch} and the \textit{Coarse grained branch}.

The \textit{Fine detailed branch} is designed to recover fine spatial details that are often lost during the encoder’s downsampling process. Specifically, the \textit{Fine detailed branch} employs transposed convolution for learnable upsampling, enabling the model to directly learn how to reconstruct fine geometric features from the training data. In the early upsampling stages, the upsampled features are concatenated with low-level feature maps from the encoder via skip connections at the exact spatial resolution, thereby preserving spatial continuity and local context. After concatenation, the fused features are further refined by an additional convolutional layer, which enhances the extraction of combined features and enriches the spatial representation. Consequently, the \textit{Fine detailed branch} plays an essential role in reconstructing object boundaries, ensuring sharpness and precision in the segmentation results.

In contrast, the \textit{Coarse grained branch} focuses on preserving the overall structure and spatial continuity during the resolution recovery process. This branch first applies a 1×1 convolution to reduce computational cost and adjust the channel of the input features, followed by bilinear interpolation - a parameter-free interpolation method that enlarges feature maps by a factor of 2× without producing the checkerboard artifacts commonly observed in transposed convolution. As a result, the \textit{Coarse grained branch} provides stable feature representations that maintain the global layout of objects, serving as a structural foundation to be integrated with the detailed features from the \textit{Fine detailed branch}.

\begin{figure}
  \centering
  \begin{subfigure}{\columnwidth}
    \centering
      \includegraphics[width=0.9\textwidth]{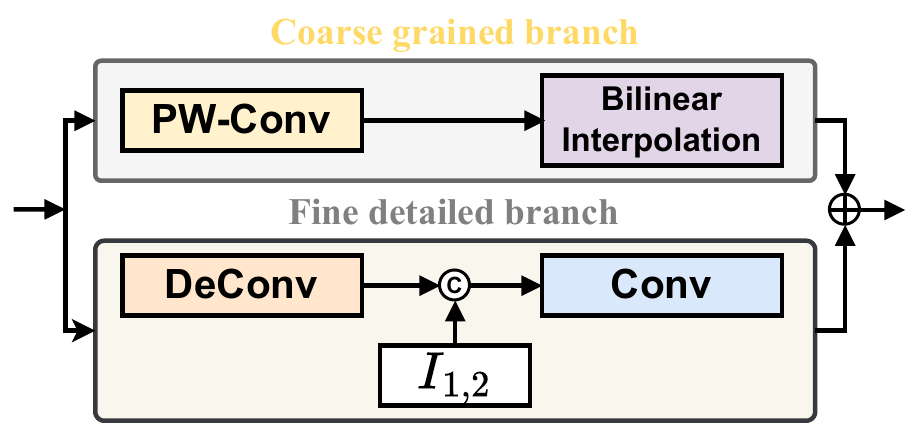}
      \caption{DBU \textit{w/ skip connection}}
      \label{dual_decoder-1}
  \end{subfigure}
  \hspace{0.057\textwidth} 
  \begin{subfigure}{\columnwidth}
  \centering
      \includegraphics[width=0.9\textwidth]{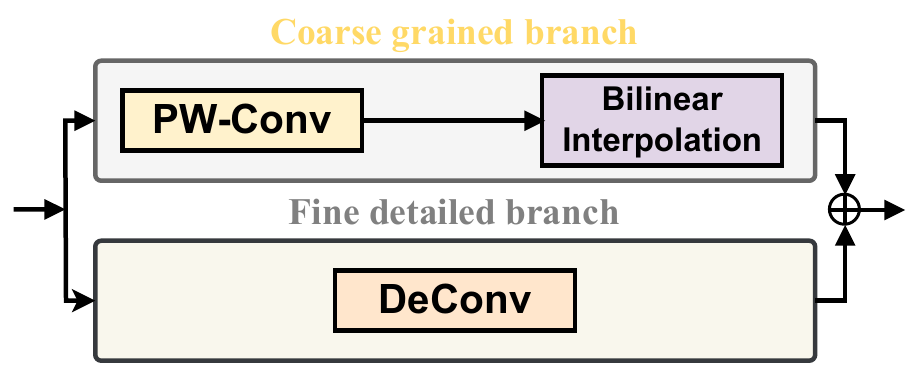}
      \caption{DBU \textit{w/o skip connection}}
      \label{dual_decoder-2}
  \end{subfigure}
  \caption{\textbf{Illustration of the proposed Dual Branch Upsampling Block (DBU).} It consists of two complementary upsampling paths: the \textit{Fine detailed branch} and the \textit{Coarse grained branch}.}
  \label{dual_decoder}
\end{figure}

The outputs from the two branches are fused through element-wise addition, allowing the model to jointly leverage the fine-detail learning capability of the transposed convolution branch and the spatial stability of the bilinear interpolation branch. This combination improves boundary accuracy and maintains a consistent spatial structure in the final segmentation results. The detailed design of the DBU is illustrated in Figure \ref{dual_decoder}, where Figure \ref{dual_decoder-1} shows the case in which the DBU incorporates downsampled features through skip connections, and Figure \ref{dual_decoder-2} depicts the configuration when the block operates independently without skip connections.

\subsection{Training strategies}

During the training process, all input images are resized from their original resolution of 1280×720 to 640×384 to ensure computational efficiency and facilitate a fair comparison with previous works that follow the same setting. To enhance generalization, we apply photometric augmentations (random hue, saturation, and value shifts) and geometric transformations (random translation, cropping, and horizontal flipping) to improve robustness and spatial diversity. We adopt the AdamW optimizer \cite{adamw} and the learning rate follows a cosine annealing schedule, gradually reducing during 100 epochs. The batch size is set to 16.

To address challenges in pixel-wise classification for autonomous driving-particularly class imbalance and fine-structure sensitivity-we design a hybrid loss combining Focal Loss~\cite{focal} and Tversky Loss~\cite{tversky}. Each loss is applied independently to the drivable-area and lane-segmentation outputs, ensuring task-specific optimization. \textbf{Focal Loss} focuses on hard-to-classify pixels by down-weighting easy examples using a modulating factor $(1-\hat{p}_i(c))^\gamma$, which is effective in mitigating severe class imbalance where background pixels dominate. \textbf{Tversky Loss} extends Dice Loss \cite{dice} by introducing weighting factors $\alpha$ and $\beta$ to balance false positives and false negatives, which is crucial for thin and elongated lane structures. The total objective is expressed as:

\begin{equation}
    \mathcal{L}_\text{total} = \mathcal{L}_\text{drivable area} + \mathcal{L}_\text{lane}
\end{equation}

Empirically, we set $\alpha$ = 0.7, $\beta$ = 0.3 for drivable-area segmentation, and $\alpha$ = 0.9, $\beta$ = 0.1 for lane segmentation, while using $\alpha_t$ = 0.25 and $\gamma$ = 2 in the Focal Loss formulation. These hyperparameters are chosen to emphasize recall over precision, ensuring safety-critical sensitivity to small or narrow structures such as lane markings. All experiments are conducted on an NVIDIA GTX 4090 GPU, with the model trained jointly for both tasks in an end-to-end multi-task setting.

\section{Experimental}


\subsection{Dataset and evaluation metrics}

The BDD100K dataset is a large-scale, diverse dataset of driving scenes developed for research in autonomous driving perception. It comprises 100,000 video clips collected from over 50,000 driving sessions across multiple regions in the United States, encompassing a wide range of environments, weather conditions, and times of day. The dataset is divided into training, validation, and test subsets containing 70K, 10K, and 20K images, respectively. Following prior studies~\cite{twin,twinplus,interactive,dffm}, since the ground-truth annotations for the test set are not publicly available, all evaluations in this work are conducted on the validation set consisting of 10,000 images. Owing to its large scale and diverse environmental conditions, BDD100K serves as a comprehensive benchmark for evaluating segmentation models under realistic driving conditions.

\begin{table}[!h]
\caption{Throughput complexity comparison among configurations at various batch sizes}
\label{tab:fps}
\resizebox{\columnwidth}{!}{
\begin{tabular}{lccccc}
\toprule
\multirow{2}{*}{\textbf{Config}} & \multicolumn{3}{c}{\textbf{FPS w/ batch size} $\uparrow$} & \multirow{2}{*}{\textbf{Params}} & \multirow{2}{*}{\begin{tabular}[c]{@{}c@{}}\textbf{FLOPs}\\ \small{batch=1}\end{tabular}} \\ \cmidrule{2-4}
                                 & \textbf{1}         & \textbf{4}        & \textbf{16}        &                                                 &                                                                                                \\ \midrule
Tiny                             & 83$_{\pm{\text{0.89}}}$                 & 335$_{\pm{\text{3.37}}}$               & 1302$_{\pm{\text{12.71}}}$               & 0.10M                                          & 1.08G                                                                                          \\[2pt]
Base                            & 67$_{\pm{\text{0.47}}}$                 & 255$_{\pm{\text{2.68}}}$               & 724$_{\pm{\text{0.94}}}$                 & 0.43M                                           & 3.95G                                                                                          \\[2pt]
Large                             & 56$_{\pm{\text{0.51}}}$                  & 218$_{\pm{\text{2.42}}}$                & 301$_{\pm{\text{0.64}}}$                 & 1.50M                                           & 14.25G                                                                                         \\ \bottomrule
\end{tabular}%
}
\end{table}

For the segmentation evaluation, consistent with prior works~\cite{twin,twinplus,interactive,dffm}, the drivable-area segmentation performance is quantified using the mean Intersection over Union (mIoU) metric. For the lane segmentation task, both Accuracy (Acc) and Intersection over Union (IoU) are employed to provide a comprehensive assessment. However, due to the substantial class imbalance between lane markings and the background, a balanced accuracy \cite{yolom,twinplus} metric is additionally adopted to yield a more reliable evaluation of model performance. To measure the computational efficiency of our approach, we adopt two commonly used indicators: the number of parameters and the FLOPs count. Following standard practice in recent studies \cite{twinplus,resnet,espnet}, FLOPs are defined as the total number of multiplication and addition operations required during inference.

\begin{table}[]
\caption{Comparison of per-epoch training time across TwinLiteNet, TwinLiteNet$^+$, and TwinMixing.}
\label{tab:training_time}
\resizebox{\columnwidth}{!}{
\begin{tabular}{lccc}
\toprule
                         & \textbf{Parameters} & \textbf{FLOPs} & \textbf{Training times} \\ \midrule
TwinLiteNet              & 0.44M      & 3.9G & 567.1s         \\
TwinLiteNet$^+_\text{Medium}$ &   0.48M      & 4.63G & 799.8s       \\
TwinMixing$_\text{base}$     & 0.43M      & 3.95G  & 1070.2s    \\ \bottomrule   
\end{tabular}%
}
\end{table}

\begin{table*}[!t]
\caption{Quantitative comparison of models on the drivable-area and lane-segmentation tasks. Results are reported as mIoU (\%) for drivable area segmentation, and as Acc (\%) and IoU (\%) for lane segmentation, together with model complexity measured by FLOPs and parameter count.}
  \centering
  \setlength{\tabcolsep}{8.5pt}
\label{tab:dall}
\begin{tabular}{lcccccc}
\toprule
\multirow{2}{*}{\textbf{Model}}                                                                     & \textbf{Drivable area segmentation}        &  & \multicolumn{2}{c}{\textbf{Lane segmentation}}             & \multirow{2}{*}{\textbf{FLOPs} } & \multirow{2}{*}{\textbf{Parameters} } \\ \cmidrule{2-2} \cmidrule{4-5}
                                                                                           & \textbf{mIoU (\%)} &  & \textbf{Acc (\%)} & \textbf{IoU (\%)} &                                     &                                      \\ \midrule
DeepLabV3+ \cite{deeplab}                                 & 90.9                 &  & {--}                     & 29.8                & 30.7G                               & 15.4M                                \\
SegFormer \cite{segformer} 				   & 92.3$^{\textcolor{red}{\textbf{4}}}$                 &  & {--}                     & 31.7                & 12.1G                               & 7.2M	                           \\
R-CNNP \cite{yolop}                                      & 90.2                 &  & {--}                     & 24.0                & {--}                                & {--}                             	   \\
YOLOP \cite{yolop}                                    & 91.6                 &  & {--}                     & 26.5                & 8.11G                               & 5.53M                                \\
YOLOv8 (multi-seg) \cite{yoloveight}                           & 84.2                 &  & 81.7$^{\textcolor{red}{\textbf{3}}}$                     & 24.3                & {--}                                & {--}                                 \\
Sparse U-PDP \cite{sparse} \textcolor{gray}                                & 91.5                 &  & {--}                     & 31.2                & {--}                                & {--}                                 \\
BILane \cite{BILane}                                   & 91.2                  &  & {--}                     & 31.3                 & {--}                                & 1.4M                                \\
EdgeUNet \cite{EdgeUNet}                                   & 89.9                  &  & {--}                     & --                 & {--}                                & --                                \\
MobiP \cite{mobip}                                   & 90.3                  &  & {--}                     & 31.2                 & 3.6G                                & 5.8M                                \\
TwinLiteNet \cite{twin}                                   & 91.3                 &  & 77.8                     & 31.1                & 3.9G                                & 0.44M                                \\\midrule
 IALaneNet${_{\footnotesize{\text{ResNet-18}}}}$ \cite{interactive}     		   & 90.5                &  & {--}                     & 30.4               & 89.83G                              & 17.05M                               \\
IALaneNet${_{\footnotesize{\text{ResNet-34}}}}$ \cite{interactive}     		   & 90.6                &  & {--}                     & 30.5               & 139.46G                             & 27.16M                               \\
IALaneNet${_{\footnotesize{\text{ConvNeXt-tiny}}}}$ \cite{interactive} 		   & 91.3                &  & {--}                     & 31.5               & 96.52G                              & 18.35M                               \\
IALaneNet${_{\footnotesize{\text{ConvNeXt-small}}}}$ \cite{interactive}  & 91.7                                  &  & {--}                     & 32.5$^{\textcolor{red}{\textbf{3}}}$               & 200.07G$^{\textcolor{red}{\textbf{6}}}$                             & 39.97M$^{\textcolor{red}{\textbf{6}}}$       
 \\ \midrule

 DFFM$_{\text{Nano}}$ \cite{dffm}   & 88.7 & & -- & 25.3 &   0.72G & 0.03M \\ 
  DFFM$_{\text{Small}}$ \cite{dffm}   & 90.8 & & -- & 29.3 &   2.56G & 0.13M \\ 
   DFFM$_{\text{Mdedium}}$ \cite{dffm}   & 92.0 & & -- & 31.6  &    10.17G & 0.5M \\ 
    DFFM$_{\text{Large}}$ \cite{dffm}   & 92.1$^{\textcolor{red}{\textbf{5}}}$ & & -- &  32.1$^{\textcolor{red}{\textbf{5}}}$ &   39.79G$^{\textcolor{red}{\textbf{5}}}$ & 2.2M$^{\textcolor{red}{\textbf{5}}}$ \\ \midrule

  TwinLiteNet$^+_{\text{Nano}}$ \cite{twinplus}                                                    & 87.3                 &  & 70.2                     & 23.3                & 0.57G                      & 0.03M                                 \\
 TwinLiteNet$^+_{\text{Small}}$ \cite{twinplus}                                                  & 90.6                 &  & 75.8                     & 29.3                & 1.40G                                & 0.12M                                \\
 TwinLiteNet$^+_{\text{Medium}}$ \cite{twinplus}                                                 & 92.0                 &  & 79.1$^{\textcolor{red}{\textbf{5}}}$                     & 32.3$^{\textcolor{red}{\textbf{4}}}$                & \underline{4.63G}                               & \underline{0.48M}                                \\
 TwinLiteNet$^+_{\text{Large}}$ \cite{twinplus}                                                  & \textbf{92.9}        &  & \underline{81.9}            & \textbf{34.2}       & 17.58G$^{\textcolor{red}{\textbf{4}}}$                              & 1.94M$^{\textcolor{red}{\textbf{4}}}$  \\ \bottomrule
  \textit{\textbf{Our proposed}}    &     &  &    &   &     &      
 \\ \toprule
 TwinMixing$_\text{tiny}$ & 91.1 && 76.6 & 29.8 & 1.08G & 0.10M \\
 TwinMixing$_\text{base}$ & 92.4$^{\textcolor{red}{\textbf{3}}}$ && 80.7$^{\textcolor{red}{\textbf{4}}}$ & \underline{33.2} & \textbf{3.95G} & \textbf{0.43M} \\
 TwinMixing$_\text{large}$ & \underline{92.8} && \textbf{82.4} & \textbf{34.2} & 14.25G$^{\textcolor{red}{\textbf{3}}}$ & 1.50M$^{\textcolor{red}{\textbf{3}}}$ \\
 \bottomrule
\end{tabular}
\vspace{2mm} 
\parbox{0.9\textwidth}{\footnotesize
The best and second-best results are marked in \textbf{bold} and \underline{underline}, while the top 3rd, 4th, and 5th ranked models are indicated with $^{\textcolor{red}{\textbf{3}}}$, $^{\textcolor{red}{\textbf{4}}}$, and $^{\textcolor{red}{\textbf{5}}}$, respectively. Note that the ranking of FLOPs and Parameters is based only on the models with the highest combined mIoU and IoU scores, rather than all models in the table.}
\end{table*}

\subsection{Main results}

\subsubsection{Inference throughput and model complexity}

The runtime characteristics and scalability of the TwinMixing family are summarized in Table~\ref{tab:fps}, detailing an evaluation of the \textit{tiny}, \textit{base}, and \textit{large} configurations at batch sizes of {1, 4, 16}. Inference FPS is computed over 500 independent runs and reported as the mean value with standard deviation. The results reveal a clear and distinct trade-off between computational complexity and inference latency, demonstrating effective throughput scaling across the architecture variants. 

The model family spans from TwinMixing$_\text{tiny}$, tailored for ultra-lightweight and edge deployments, to TwinMixing$_\text{large}$, which targets high-capacity scenarios. The TwinMixing$_\text{tiny}$ variant demonstrates excellent hardware utilization, achieving 83 FPS at batch size 1 and scaling near-linearly to 1302 FPS at batch size 16. This indicates the model is predominantly compute-bound and highly optimized for resource-constrained applications. For a balanced latency–capacity profile, TwinMixing$_\text{base}$ provides substantially greater model capacity than the \textit{tiny} variant while maintaining robust real-time throughput, delivering 67 FPS with a batch size of 1 and 724 FPS with a batch size of 16. The \textit{large} variant is designed for high-end GPU deployments where model capacity and accuracy are prioritized. Despite incurring the highest computational cost, it sustains real-time operation at 56 FPS with a batch size of 1 and scales effectively to 301 FPS with a batch size of 16. These results confirm the effective scalability of the TwinMixing architecture, whose hierarchical variants offer favorable throughput-complexity trade-offs suitable for diverse computational budgets.

\subsubsection{Training time comparison}

We compare the training time of TwinLiteNet \cite{twin}, TwinLiteNet$^+$ \cite{twinplus}, and TwinMixing on an RTX 4090 GPU. Table~\ref{tab:training_time} highlights a clear difference between theoretical computational efficiency (FLOPs and parameter count) and practical training cost (time per epoch) across the three models. Although TwinMixing achieves the lowest FLOPs (3.9G) and a compact parameter size (0.44M) compared with the TwinLiteNet variants, its training time per epoch increases substantially to 1070,2s.
This gap is primarily attributed to differences in the core operators and the resulting computational graph complexity. Specifically, TwinLiteNet relies on dilated convolutions as its primary feature-extraction operator, while TwinLiteNet$^+$ adopts dilated depthwise convolutions to improve efficiency. The complex design of this unit results in a more intricate computation graph, increasing the cost of backpropagation and gradient computation. Moreover, grouped convolution and channel shuffle are introduced to improve cross-channel interaction without notably increasing FLOPs; they can introduce additional overhead from memory access and frequent tensor layout transformations on the GPU, further prolonging training. Finally, the Dual Branch Upsampling (DBU) decoder, which processes two parallel branches to recover fine details, increases intermediate activation volume and thus adds training-time overhead.

Despite its more complex architecture, which aims to enhance feature extraction and reconstruction, TwinMixing maintains low FLOPs and parameter count through width scaling, carefully tuning channel width across layers. This design choice reduces both the number of parameters and FLOPs relative to the compared baselines, thereby improving deployment efficiency, at the expense of higher training cost due to multi-branch computation and memory-related overhead.

\subsubsection{Quantitative results}

The compared methods include both general-purpose segmentation networks and multi-task perception models, ranging from standard baselines \cite{deeplab,segformer,yolop,sparse} to recent lightweight architectures \cite{twin,mobip,BILane}. We also include scalable models with multiple configurations \cite{dffm,interactive,twinplus}, which are analogous to our TwinMixing for a fair comparison across different scales. The quantitative results of all models are summarized in Table~\ref{tab:dall}.

On the BDD100K dataset, the proposed TwinMixing achieves high segmentation accuracy while maintaining low computational cost.
For the lane segmentation task, TwinMixing$_\text{large}$ reaches 82.4\% accuracy and 34.2\% IoU, outperforming competitive models such as TwinLiteNet$^+_\text{Large}$, DFFM$_\text{Large}$, and IALaneNet$_\text{ConvNeXt-small}$. Despite higher accuracy, our model requires significantly fewer resources, highlighting the effectiveness of the proposed architecture in modeling fine-grained structural cues with minimal overhead. For the drivable area segmentation task, TwinMixing$_\text{large}$ achieves 92.8\% mIoU, which is only 0.1\% lower than TwinLiteNet$^+_\text{Large}$, but with a substantially lower computational burden, saving 3.33 GFLOPs and 0.44M parameters.
This demonstrates that the large configuration of TwinMixing offers a superior balance between accuracy and efficiency across both segmentation tasks compared to prior multi-task networks such as TwinLiteNet$^+$, DFFM, and IALaneNet.

\begin{figure*}
    \centering
    \includegraphics[width=\linewidth]{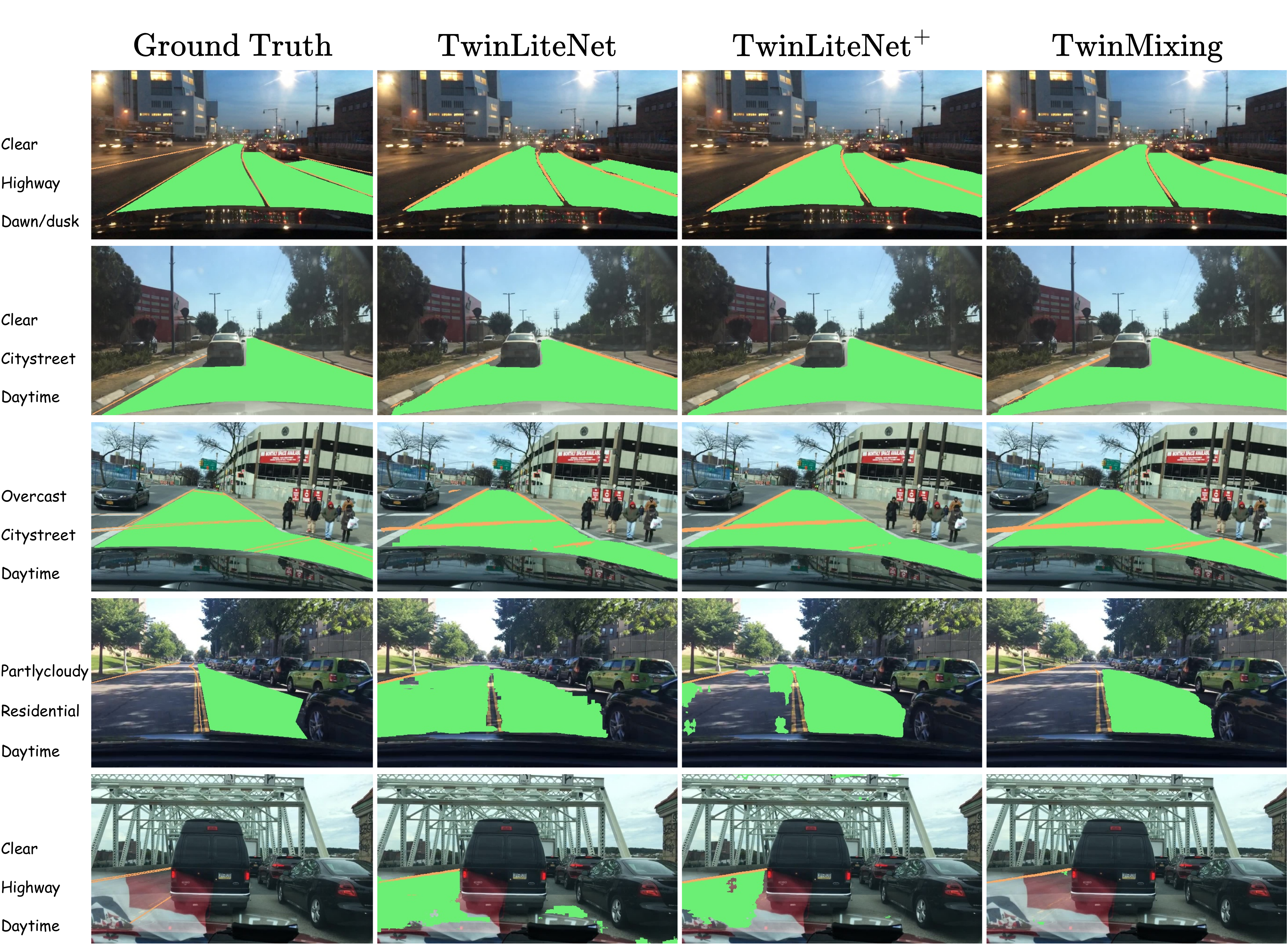}
    \caption{Qualitative comparison of segmentation results under normal driving conditions.}
    \label{fig:normalconds}
\end{figure*}

\begin{figure*}
    \centering
    \includegraphics[width=1.\linewidth]{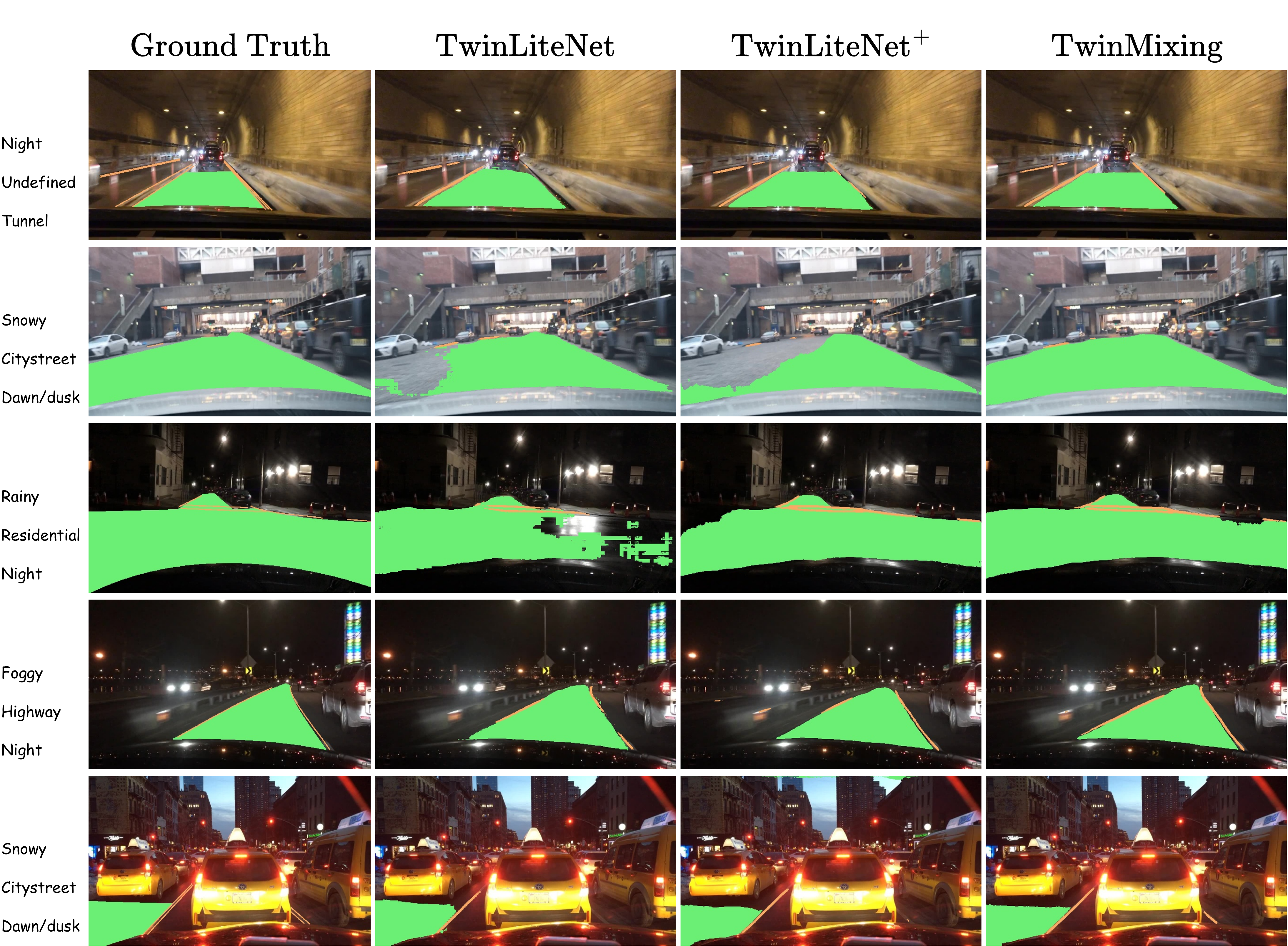}
    \caption{Qualitative comparison of segmentation results under challenging driving conditions.}
    \label{fig:challengeconds}
\end{figure*}

\begin{figure*}
    \centering
    \includegraphics[width=\linewidth]{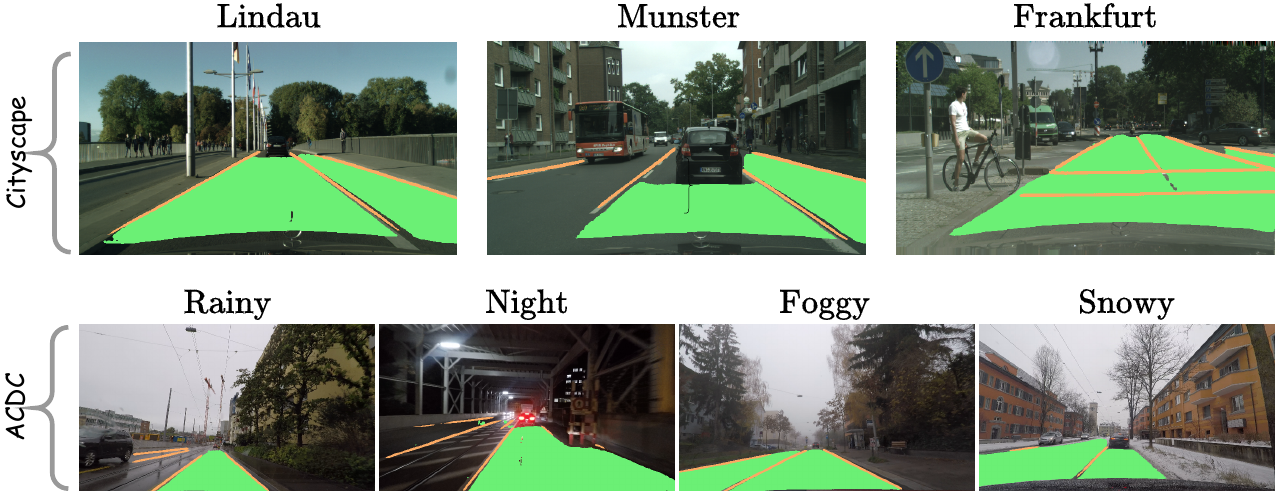}
    \caption{Qualitative visualization of cross-dataset generalization.}
    \label{fig:across}
\end{figure*}

Among the top five models with the highest overall performance, measured by the sum of mIoU for drivable area segmentation and IoU for lane segmentation, TwinMixing$_\text{base}$ demonstrates the best computational efficiency, requiring only 3.95 GFLOPs and 0.43M parameters while maintaining competitive accuracy (92.4\% mIoU, 33.2\% IoU). Its performance ranks just below TwinMixing$_\text{large}$ and TwinLiteNet$^+_\text{large}$, yet with significantly lower computational cost, highlighting its strong accuracy–efficiency balance. Furthermore, the TwinMixing$_\text{tiny}$ variant, tailored for ultra-lightweight embedded deployment, achieves 91.1\% mIoU and 29.8\% IoU with only 1.08 GFLOPs and 0.10M parameters, demonstrating excellent scalability under limited computational budgets.

These results clearly demonstrate that TwinMixing achieves an outstanding balance between segmentation accuracy and computational cost compared to existing multi-task segmentation models. As illustrated in Figure 2, the proposed models consistently lie on the optimal region of the accuracy–efficiency curve, showing favorable trade-offs across different configurations. This confirms the scalability of TwinMixing, whose hierarchical variants (\textit{tiny}, \textit{base}, \textit{large}) maintain competitive performance under varying computational budgets.

\subsection{Qualitative results}

\subsubsection{Qualitative analysis on BDD100K}

For a comprehensive evaluation, we also evaluate TwinMi-
xing$_\text{base}$ configurations with TwinLiteNet and TwinLiteNet$^+_\text{Medium}$ using their public pretrained checkpoints on BDD100K~\cite{bdd} with standardized configurations. These two models have the same parameters as TwinMixing, making them appropriate baselines for comparison. Evaluation is conducted on the BDD100K validation split, with subsets stratified by time of day, scene type, and weather. Figure~\ref{fig:normalconds} presents a visual comparison across highway, city street, and residential settings with benign weather in daylight and dawn/dusk conditions. TwinMixing produces cleaner drivable-area masks with tighter lane boundaries than TwinLiteNet and TwinLiteNet$^+$. In residential and highway scenes under daytime lighting, it preserves lane geometry and road–horizon alignment, reducing spillover into non-drivable regions frequently observed in the baselines. The model also remains reliable under illumination changes on highway scenes at dawn/dusk and on city-street scenes in both clear and overcast daytime conditions. TwinMixing maintains accurate lane delineation while suppressing spurious activations near vehicles and roadside structures.

Beyond benign settings, we qualitatively assess robustness in adverse environments (Fig.~\ref{fig:challengeconds}), spanning night time lighting with city street, residential, highway, and tunnel scenes under snow, rain, and fog conditions. TwinMixing offers cleaner drivable-area estimates with tighter lane demarcation, outperforming TwinLiteNet and TwinLiteNet$^+$. In a city street scene under snowy weather conditions at dawn, TwinMixing captures lane boundaries missed by the baselines and provides superior delineation. Across tunnel, residential, and highway scenes under these conditions, TwinMixing maintains stable drivable-area masks and sharper lane boundaries; in particular, on residential scenes at night and in rain, it yields a cleaner drivable-area mask than TwinLiteNet. These observations are consistent with the qualitative improvements in boundary quality and the reduction in false positives.

\subsubsection{Cross-dataset generalization analysis}

Beyond in-domain evaluation on the BDD100K dataset, we further assess the cross-dataset generalization ability of TwinMixing on Cityscapes \cite{cityscape} and ACDC \cite{ACDC}. The Cityscapes dataset consists of urban scenes captured across different European cities. At the same time, ACDC focuses on diverse and challenging environmental conditions. Qualitative results of TwinMixing$^+_\text{large}$ across these datasets are visualized in Figure \ref{fig:across}. The results indicate that TwinMixing exhibits strong generalization across unseen datasets, successfully recognizing drivable areas and lane regions in scenes captured from different geographic locations and under adverse environmental conditions, including nighttime, rain, fog, and snow. Nevertheless, due to domain shifts and the absence of training on these datasets, the model does not consistently achieve optimal accuracy in specific scenarios, highlighting remaining challenges in cross-domain robustness.

\begin{table*}[t]
\centering
\setlength{\tabcolsep}{11pt}
\caption{Comparative evaluation of the TwinMixing model under multi-task and single-task learning settings, reporting segmentation accuracy and computational efficiency metrics.}
\label{tab:multitask}
\begin{tabular}{lccccc}
\toprule
\multirow{2}{*}{\textbf{Method}} & \textbf{Drivable area segmentation} &  \multicolumn{2}{c}{\textbf{Lane segmentation}}              & \multirow{2}{*}{\textbf{Parameters} } & \multirow{2}{*}{\textbf{FLOPs} } \\ \cmidrule(lr){2-2} \cmidrule(lr){3-4}
                                 & \textbf{mIoU (\%)}        &  \textbf{Acc (\%)}  & \textbf{IoU (\%)}  &                                                 &                                              \\ \midrule
\multirow{2}{*}{Single-task}     & 92.3                                &  \xmark                       & \xmark                       & 0.41M                                           & 3.50G                                       \\
                                 & \xmark                              &  81.1                         & 33.5                         & 0.41M                                           & 3.50G                                       \\ \midrule
Multi-task                       & 92.4$_{\textcolor{blue}{\uparrow\text{0.1}}}$                                & 80.7$_{\textcolor{red}{\downarrow\text{0.4}}}$                         & 33.2$_{\textcolor{red}{\downarrow\text{0.3}}}$                         & 0.43M$_{\textcolor{red}{\uparrow\text{0.02M}}}$                                           & 3.95G$_{\textcolor{red}{\uparrow\text{0.45G}}}$                                       \\ \bottomrule
\end{tabular}%
\end{table*}

\subsection{Ablation study}

\subsubsection{Multi-task and single-task models}

Table~\ref{tab:multitask} contrasts TwinMixing trained as two separate single-task models with a single multi-task model for drivable-area and lane segmentation. The multi-task configuration attains a slightly lower drivable-area mIoU 92.3\% vs. 92.4\%, matches lane accuracy 81.1\%, and achieves a marginally higher lane IoU 33.5\% vs. 33.2\%. Notably, a single multi-task model requires only 0.434\,M parameters and 3.950\,G FLOPs to produce both outputs, whereas deploying two single-task models together requires about 0.823\,M parameters and 6.998\,G FLOPs. Relative to a single-task model, the multi-task variant adds only $\sim$0.023\,M parameters and $\sim$0.451\,G FLOPs, yet replaces two models with one, delivering near-parity accuracy at roughly half the total compute and parameter budget. These results indicate that TwinMixing offers a balanced and efficient trade-off between performance and complexity for joint drivable-area and lane segmentation in multi-task settings.

\subsubsection{The results of TwinMixing in different conditions}

Table~\ref{tab:envconds} presents the performance of TwinMixing$_\text{base}$ across a range of environmental conditions in the BDD100K. We evaluate the model under favorable conditions with clear illumination and simple scene layouts—such as daytime, highway, and residential before moving to more challenging scenarios, including nighttime, rainy, and tunnel environments. The results indicate that TwinMixing performs strongly under normal conditions, maintaining high segmentation accuracy, while its performance degrades moderately under adverse environments, such as rainy and tunnel conditions. Notably, despite the challenging lighting conditions in night scenes, the model still achieves competitive results (92.5\% mIoU for drivable area and 32.9\% IoU for lane segmentation), which can be attributed to its exposure to a large number of nighttime images (~28k/70k) during training. Adverse conditions primarily affect lane segmentation, as reflective surfaces, motion blur, and low-contrast lane markings reduce model confidence. Among all, tunnel scenes pose the greatest challenge, where lane IoU drops significantly compared to the overall performance. These findings demonstrate that TwinMixing maintains robust, consistent accuracy in drivable area segmentation. In contrast, lane detection remains more sensitive to challenging conditions, such as foggy, snowy, and tunnel environments.

\begin{table}[]
\centering
\setlength{\tabcolsep}{9pt}
\caption{Performance across various environmental conditions on both task drivable area segmentation and lane segmentation}
\label{tab:envconds}
\begin{tabular}{lcc}
\toprule
\multirow{2}{*}{\textbf{\begin{tabular}[c]{@{}l@{}}Environmental\\ conditions\end{tabular}}} & \textbf{\begin{tabular}[c]{@{}c@{}}Drivable area\\ segmentation\end{tabular}} & \textbf{Lane segmentation}  \\ \cmidrule(lr){2-2} \cmidrule(l){3-3} 
                                                                                             & \textbf{mIoU (\%)}                                                  & \textbf{IoU (\%)}  \\ 
                                                                                             \midrule
Daytime                                                                                      & 93.2                                                                          & 35.1                                                                  \\
Night                                                                                        & 92.5                                                                          & 32.9                                                                  \\
Snowy                                                                                        & 91.6                                                                          & 31.6                                                                  \\
Rainy                                                                                        & 90.2                                                                          & 32.0                                                                  \\
Foggy                                                                                        & 92.4                                                                          & 29.6                                                                  \\
Highway                                                                                      & 93.3                                                                          & 33.8                                                                  \\
Residential                                                                                  & 92.8                                                                          & 34.4                                                                  \\
Tunnel                                                                                       & 89.8                                                                          & 28.9                                                                  \\ 
\midrule
\textbf{TwinMixing$_\text{base}$}                                                                                      & \textbf{92.4}                                                                          & \textbf{33.2}                                                                  \\\bottomrule
\end{tabular}%
\end{table}

\subsubsection{Ablation study on the Dual Branch Upsampling module}

To evaluate the contribution of each component within the proposed Dual Branch Upsampling (DBU) module, we conduct an ablation analysis by selectively removing the fine-detailed branch and coarse-grained branch. The results are summarized in Table~\ref{tab:ab_dbu}. When either the fine-detailed branch (Transposed Convolution–based) or the coarse-grained branch (bilinear interpolation–based) is omitted, a consistent performance degradation is observed across both drivable-area and lane segmentation tasks. Specifically, removing the fine-detailed branch results in a 0.2\% drop in mIoU and a 0.3\% reduction in IoU, indicating its importance in restoring spatial details. Similarly, eliminating the coarse-grained branch results in slightly lower accuracy and IoU, confirming its role in providing spatial smoothness and stability. These findings demonstrate that the complementary design of the two branches enables DBU to achieve a better balance between fine-grained reconstruction and smooth upsampling, thereby improving overall segmentation performance.

\begin{table}[]
\centering
\caption{Ablation of the Dual Branch Upsampling (DBU) module. Removing either branch degrades performance, confirming their complementary effect on segmentation accuracy.}
\label{tab:ab_dbu}
\begin{tabular}{lccc}
\toprule
\multirow{2}{*}{\textbf{Methods}}         & \textbf{\begin{tabular}[c]{@{}c@{}}Drivable area\\ segmentation\end{tabular}} & \multicolumn{2}{c}{\textbf{Lane segmentation}} \\ \cmidrule{2-2}  \cmidrule(l){3-4}
                                   & \textbf{mIoU (\%)}                                                            & \textbf{Acc (\%)}      & \textbf{IoU (\%)}     \\ \midrule
DBU     & 92.4  & 80.7 & 33.2                  \\[4pt]
\small{\textit{w/o fine detailed}} & 92.2$_{\textcolor{red}{\downarrow\text{0.2}}}$ & 80.3$_{\textcolor{red}{\downarrow\text{0.4}}}$ & 32.9$_{\textcolor{red}{\downarrow\text{0.3}}}$                  \\
\small{\textit{w/o coarse grained}}  & 92.3$_{\textcolor{red}{\downarrow\text{0.1}}}$ & 80.3$_{\textcolor{red}{\downarrow\text{0.4}}}$ & 33.0$_{\textcolor{red}{\downarrow\text{0.2}}}$                 \\ \bottomrule
\end{tabular}%
\end{table}

\subsubsection{Sensitivity analysis of dilation rate and group size in EPM module}

In this section, we analyze the sensitivity of the proposed model to the dilation rate and group size configurations in the EPM module. All experiments are conducted using the TwinMixing$_\text{tiny}$ configuration, trained for 50 epochs under identical settings. Different dilation-rate and group-size variants are compared against the default TwinMixing$_\text{tiny}$ configuration, which is also trained for 50 epochs to ensure a fair comparison.

To evaluate the effect of dilation rates, we replace the default multi-dilation setting in the EPM module with fixed dilation rates of 1, 4, and 16, while keeping all other components unchanged. As reported in Table \ref{tab:dilate}, the default multi-dilation configuration achieves the best performance, attaining 90.9\% mIoU for drivable-area segmentation and 29.4\% IoU for lane segmentation. In contrast, using a single fixed dilation rate consistently degrades performance, with more pronounced drops observed for lane segmentation. These results indicate that the multi-dilation design is more effective at capturing multi-scale contextual information and mitigating gridding artifacts than fixed-dilation-rate alternatives.

We further examine the sensitivity to group size in grouped convolutions. While the input and output channel dimensions determine the default group size, we override this rule and set the number of groups to \{1, 2, 4, 8, 16, 32\}. As shown in Table \ref{tab:group}, varying the group size primarily affects model complexity, as measured by the number of parameters and FLOPs. In contrast, segmentation accuracy remains stable mainly across different configurations. Smaller group sizes yield marginal improvements in accuracy at the expense of increased computation. In comparison, larger group sizes reduce the number of parameters and FLOPs with negligible performance degradation. Overall, the results indicate a clear efficiency–accuracy trade-off and confirm that TwinMixing is not highly sensitive to the specific group-size choice.

\begin{table}[]
\centering
\caption{Sensitivity analysis of dilation-rate configurations in the EPM module.}
\label{tab:dilate}
\begin{tabular}{cccc}
\toprule
\multicolumn{1}{l}{\multirow{2}{*}{\textbf{Dilated rates}}} & \textbf{\begin{tabular}[c]{@{}c@{}}Drivable area\\ segmentation\end{tabular}} & \multicolumn{2}{c}{\textbf{Lane segmentation}} \\ \cmidrule(lr){2-2}  \cmidrule(lr){3-4} 
\multicolumn{1}{l}{}                                        & \textbf{mIoU (\%)}                                                            & \textbf{Acc (\%)}      & \textbf{IoU (\%)}     \\ \midrule
   $\bigstar$                                                         & 90.9                                                                          & 76.4                   & 29.4                  \\
1                                                           & 89.6                                                                          & 75.4                   & 28.2                  \\
4                                                           & 90.2                                                                          & 74.8                   & 27.7                  \\
16                                                          & 89.8                                                                          & 74.8                   & 27.7                  \\ \bottomrule
\end{tabular}%
\vspace{2mm} 
\parbox{0.9\textwidth}{\footnotesize
$\bigstar$ indicate default settings}
\end{table}

\begin{table}[]
\centering
\caption{Sensitivity analysis of group-size configurations  in the EPM module.}
\label{tab:group}
\begin{tabular}{ccccc}
\toprule
\multicolumn{1}{l}{\multirow{2}{*}{\textbf{Groups}}} & \multicolumn{1}{l}{\multirow{2}{*}{\textbf{Parameters}}} & \multicolumn{1}{l}{\multirow{2}{*}{\textbf{FLOPs}}} & \textbf{Drivable area} & \textbf{Lane}              \\ \cmidrule(lr){4-4} \cmidrule(lr){5-5} 
\multicolumn{1}{l}{}                                 & \multicolumn{1}{l}{}                                     & \multicolumn{1}{l}{}                                & \textbf{mIoU (\%)}     & \textbf{IoU (\%)} \\  \midrule
$\bigstar$                                 & 98.2K & 1,084G                                & 90.9                                                                          & 29.4                                                        \\
1                                                    & 115.0K                                                   & 1.222G                                              & 91.3                                                                          & 30.4                                                        \\
2                                                    & 104.5K                                                   & 1.118G                                              & 91.2                                                                          & 29.5                                                        \\
4                                                    & 100.7K                                                   & 1.094G                                              & 90.8                                                                          & 29.3                                                        \\
8                                                    & 99.2K                                                    & 1.088G                                              & 90.9                                                                          & 29.3                                                        \\
16                                                   & 98.5K                                                    & 1.086G                                              & 90.9                                                                          & 29.2                                                        \\
32                                                   & 98.2K                                                    & 1.084G                                              & 90.9                                                                          & 29.3                                                        \\ \bottomrule
\end{tabular}%
\vspace{2mm} 
\parbox{0.9\textwidth}{\footnotesize
$\bigstar$ indicate default settings}
\end{table}

\subsection{Quantization and deployment}

Table \ref{tab:quant} presents the quantization performance of TwinMixing under three numerical precisions: FP32, FP16, and INT8. For the INT8 configuration, we employ Quantization-Aware Training (QAT), which integrates quantization operations directly into the training process. Instead of retraining the model from scratch, we fine-tune the pre-trained FP32 model for 10 additional epochs using QAT. As shown, quantization has a negligible impact on segmentation accuracy while significantly reducing computational cost and memory usage-a crucial advantage for real-time deployment on embedded hardware.

In particular, the FP16 configuration achieves nearly identical accuracy compared to the FP32 baseline. Transitioning to INT8 quantization introduces only a marginal performance degradation (< 1\%) while reducing model size by up to 4× and substantially lowering inference latency across various hardware platforms. These results confirm the robustness of TwinMixing to precision scaling, highlighting its suitability for efficient inference on edge and low-power devices without compromising segmentation quality.

To further evaluate the real-time inference capability of TwinMixing on embedded hardware, we measure the latency of the \textit{tiny} configuration across multiple NVIDIA Jetson platforms, including AGX Orin, Xavier, Orin Nano, and TX2. The inference is executed using TensorRT with FP16 precision to leverage hardware acceleration and optimize runtime efficiency. The results, presented in Table \ref{tab:deloy}, demonstrate that TwinMixing maintains consistently low inference latency across all devices, with 21.96 ms on AGX Orin and ~27 ms on Xavier and Orin Nano, while remaining under 60.81 ms even on the older TX2 board. These results indicate that TwinMixing is well-suited for real-time deployment on a wide range of embedded systems with varying computational capacities.

\begin{table}[]
\centering
\setlength{\tabcolsep}{9pt}
\caption{Quantization results of TwinMixing across different configurations, including FP32, FP16, and INT8. Results are presented in the format mIoU (for drivable area segmentation) / IoU (for lane segmentation).}
\label{tab:quant}
\begin{tabular}{lccc}
\toprule
\multirow{2}{*}{\textbf{Config}} & \multicolumn{3}{c}{\textbf{Performance}}                               \\ \cmidrule{2-4} 
               & \textbf{FP32} & \textbf{FP16} &  \textbf{INT8 (QAT)} \\ \midrule
Tiny  & 91.1 / 29.8     & 91.1 / 29.8          & 90.3 / 29.2         \\
Base  & 92.4 / 33.2     & 92.4 / 33.2         & 92.2 / 32.7         \\
Large & 92.8 / 34.2     & 92.8 / 34.2         & 92.6 / 33.9         \\ \bottomrule
\end{tabular}%
\end{table}

\section{Discussion and Conclusion}

\subsection{Limitations}

\begin{figure}
    \centering
    \includegraphics[width=\linewidth]{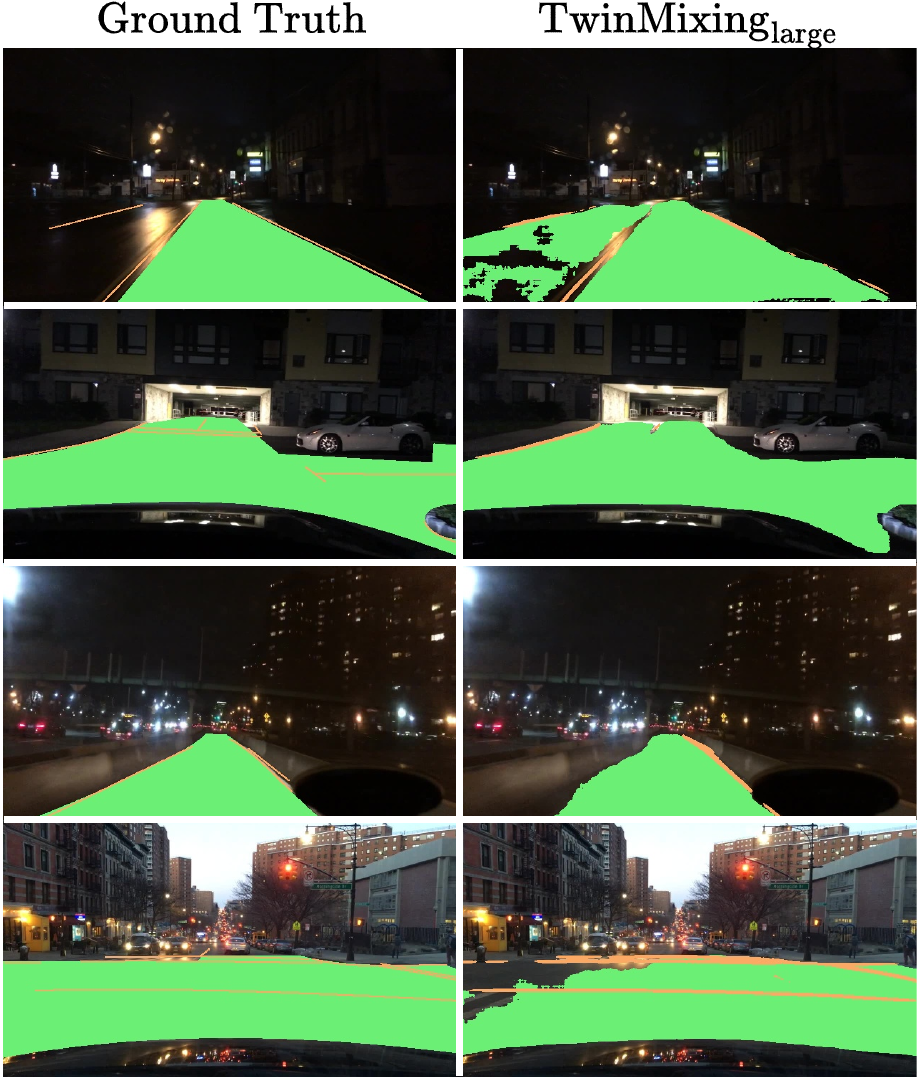}
    \caption{Visualization of challenging failure cases for TwinMixing$_\text{large}$.}
    \label{fig:fail}
\end{figure}

Although TwinMixing demonstrates strong overall performance, several limitations remain that merit further investigation. First, despite its robustness across diverse driving environments, the model’s drivable area and lane segmentation accuracy degrade under adverse conditions such as snow, rain, or tunnels, as shown in Table \ref{tab:envconds}. This suggests that TwinMixing's reliability under varying illumination and visibility could be enhanced through more advanced data augmentation or domain adaptation techniques. Second, while the proposed Efficient Pyramid Mixing (EPM) and Dual-Branch Upsampling (DBU) modules achieve an excellent balance between accuracy and efficiency, their configurations still depend on manually tuned architectural hyperparameters (e.g., dilation rates, grouping factors, and repetition depth). Incorporating neural architecture search (NAS) or adaptive parameterization strategies could further optimize performance and generalizability. Finally, TwinMixing currently focuses solely on segmentation tasks. Extending it to support additional perception tasks such as object detection, depth estimation, or panoptic segmentation could form a unified perception backbone for broader autonomous driving applications.

We also additionally present representative failure cases of TwinMixing (large configuration) in Figure \ref{fig:fail}. While the model demonstrates competitive performance across various conditions in Figure \ref{fig:normalconds},\ref{fig:challengeconds}, it still struggles in low-light nighttime scenes and under adverse weather conditions such as rain or snow (Figure \ref{fig:fail}). In these scenarios, strong illumination contrast, reflections on wet road surfaces, and increased visual noise hinder effective feature extraction, resulting in inaccurate drivable-area boundaries and incomplete or fragmented lane segmentation. These observations highlight the model’s current limitations under extreme environmental conditions and suggest that further robustness enhancements, such as illumination-aware training strategies or multimodal cues, could be beneficial.

\begin{table}[]
\centering
\caption{Inference latency (in milliseconds) of TwinMixing$_\text{tiny}$ on various NVIDIA Jetson devices, reported as the mean and standard deviation over 500 independent runs.}
\label{tab:deloy}
\begin{tabular}{lcccc}
\toprule
\textbf{Device} & AGX Orin & Xavier & Orin Nano & TX2 \\ \midrule
\textbf{Latency} & 21.96$_{\pm{\text{0.23}}}$            & 26.78$_{\pm{\text{0.38}}}$                 & 27.53$_{\pm{\text{0.08}}}$             &    60.81$_{\pm{\text{0.13}}}$          \\ \bottomrule
\end{tabular}%
\end{table}

\subsection{Conclusion}

In this work, we present TwinMixing, a shuffle-aware, lightweight multi-task segmentation model designed explicitly for drivable area and lane segmentation in autonomous driving. The proposed Efficient Pyramid Mixing (EPM) module enhances multi-scale feature extraction. At the same time, the Dual Branch Upsampling (DBU) block improves decoding stability by combining fine-detailed and coarse-grained spatial reconstruction. Comprehensive experiments on the BDD100K dataset demonstrate that TwinMixing achieves a superior trade-off between accuracy and efficiency, outperforming state-of-the-art lightweight models such as TwinLiteNet+ and DFFM while requiring substantially fewer parameters and FLOPs. Moreover, its consistent real-time inference speed across NVIDIA Jetson devices confirms its potential for embedded and edge deployment. Future work will focus on improving model robustness under extreme weather and lighting conditions, exploring self-supervised pretraining for better generalization, and extending TwinMixing to broader panoptic perception tasks in autonomous systems.



\bibliographystyle{IEEEtran}
\bibliography{ref}

\end{document}